\def\year{2019}\relax
\documentclass[letterpaper]{article} 
\usepackage{aaai19}  
\usepackage{times}  
\usepackage{helvet}  
\usepackage{courier}  
\usepackage{url}  
\usepackage{graphicx}  

\usepackage{epsfig}
\usepackage{amsmath}
\usepackage{amssymb}
\usepackage{color}
\usepackage{comment}
\usepackage{algorithm}
\usepackage{algorithmic}

\usepackage{enumitem}
\usepackage{amsfonts}
\usepackage{extarrows}
\usepackage{multirow}
\usepackage{rotating}

\begin{document}
%
\title{Network Transplanting (extended abstract)}
\author{Quanshi Zhang$^{a}$, Yu Yang$^{b}$, Qian Yu$^{c}$ and Ying Nian Wu$^{b}$\\
$^{a}$Shanghai Jiao Tong University,\quad$^{b}$University of California, Los Angeles,\quad$^{c}$University of California, Berkeley}
\maketitle

\section{Introduction}

\footnote[1]{Quanshi Zhang is the corresponding author with the John Hopcroft Center and the MoE Key Lab of Artificial Intelligence, AI Institute, Shanghai Jiao Tong University.}Besides end-to-end learning a black-box neural network, in this paper, we propose a new deep-learning methodology, \emph{i.e.} network transplanting. Instead of learning from scratch, network transplanting aims to merge several convolutional networks that are pre-trained for different categories and tasks to build a generic, distributed neural network.

Network transplanting is of special values in both theory and practice. We briefly introduce key deep-learning problems that network transplanting deals with as follows.

\subsection{Future potential of learning a universal net}
\label{sec:universal}

Instead of learning different networks for different applications, building a universal net with a compact structure for various categories and tasks is one of ultimate objectives of AI. In spite of the gap between current algorithms and the target of learning a huge universal net, it is still meaningful for scientific explorations along this direction. Here, we list key issues of learning a universal net, which are not commonly discussed in the current literature of deep learning.

\noindent
$\bullet\;$\textbf{The start-up cost \emph{w.r.t.} sample collection} is also important, besides the total number of training annotations. Traditional methods usually require people to simultaneously prepare training samples for all pairs of categories and tasks before the learning begins. However, it is usually unaffordable, when there is a large number of categories and tasks. In comparison, our method enables a neural network to sequentially absorb network modules of different categories one-by-one, so the algorithm can start without all data.

\noindent
$\bullet\;$\textbf{Massive distributed learning \& weak centralized learning:} Distributing the massive computation of learning the network into local computation centers all over the world is of great practical values. There exist numerous networks locally pre-trained for specific tasks and categories in the world. Centralized network transplanting physically merges these networks into a compact universal net with a few or even without any training samples.

\noindent
$\bullet\;$\textbf{Delivering models or data:} Our delivering pre-trained networks to the computation center is usually much cheaper than collecting and sending raw training data in practice.

\begin{figure*}[t]
\centering
\includegraphics[width=0.9\linewidth]{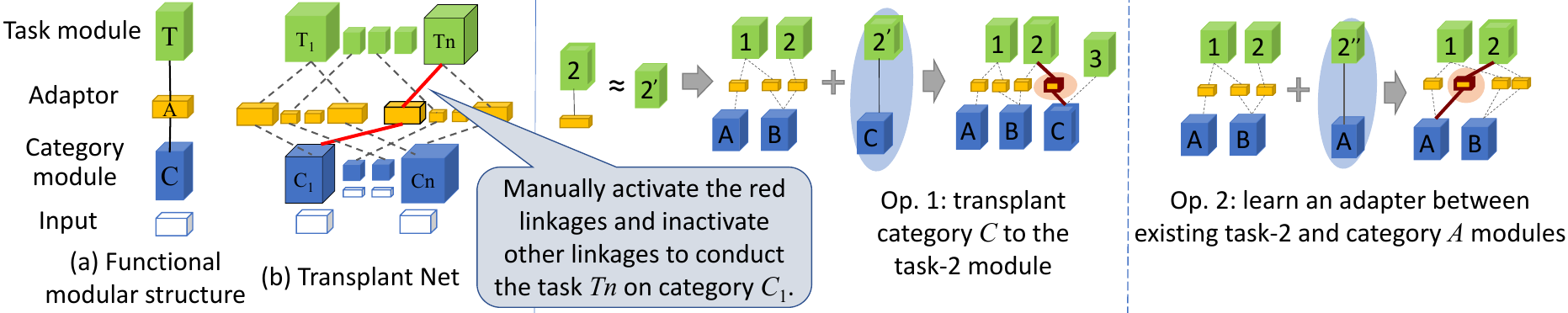}
\vspace{-10pt}
\caption{Building a transplant net. We propose a theoretical solution to incrementally merging category modules from teacher nets into a transplant (student) net with a few or without sample annotations. The transplant net has an interpretable, modular structure. A category module, \emph{e.g.} a cat module, provides cat features to different task modules. A task module, \emph{e.g.} a segmentation module, serves for various categories. We show two typical operations to learn transplant nets. Blue ellipses show modules in teacher nets used for transplant. Red ellipses indicate new modules added to the transplant net. Unrelated adapters in each step are omitted for clarity.}
\label{fig:problem}
\vspace{-10pt}
\end{figure*}

\noindent
$\bullet\;$\textbf{Middle-to-end semantic manipulation for application:} How to efficiently organize and use the knowledge in the net is also a crucial problem. We use different modules in the network to encode knowledge of different categories and that of different tasks. Like building LEGO blocks, people can manually connect a category module and a task module to accomplish a certain application (see Fig.~\ref{fig:problem}(left)).

\begin{table*}
\begin{center}
\resizebox{\linewidth}{!}{
\begin{tabular}{p{2.8cm}|p{3.5cm}p{4.5cm}cccc}
& \multicolumn{1}{|c}{Annotation cost} & \multicolumn{1}{c}{Sample preparation} & \!\!Interpretability\!\! & Catastrophic forgetting & \!\!{\small Modular manipulate}\!\! & Optimization\\
\hline
Directly learning a multi-task net & Massive & Simultaneously prepare samples for all tasks and categories & Low & -- & Not support & back prop.\\
\hline
Transfer- / meta- / continual-learning & Some support weakly-supervised learning & Some learn a category/task after another & Usually low & \!\!Most algorithmically alleviate\!\! & No support & back prop.\\
\hline
Transplanting & A few or w/o annotations & Learns a category after another & High & Physically avoid & Support & \!\!back-back prop.\!\!\\
\hline
\end{tabular}}
\vspace{-10pt}
\caption{Comparison between network transplanting and other studies. Note that this table can only summarize mainstreams in different research directions considering the huge research diversity.\vspace{-20pt}}
\label{tab:diff}
\end{center}
\end{table*}

\subsection{Task of network transplanting}

To solve above issues, we propose network transplanting, \emph{i.e.} building a generic model by gradually absorbing networks locally pre-trained for specific categories and tasks. We design an interpretable modular structure for a target network, namely a \textit{transplant net}, where each module is functionally meaningful. As shown in Fig.~\ref{fig:problem}(left), the transplant net consists of three types of modules, \emph{i.e.} category modules, task modules, and adapters. Each category module extracts general features for a specific category (\emph{e.g.} the dog). Each task module is learned for a certain task (\emph{e.g.} classification or segmentation) and is shared by different categories. Each adapter projects output features of a category module to the input space of a task module. Each category/task module is shared by multiple tasks/categories.

We can learn an initial transplant net with very few tasks and categories in the scenario of traditional multi-task/category learning. Then, we gradually grow the transplant net to deal with more categories and tasks via network transplanting. Network transplanting can be conducted with or without human annotations as additional supervision. We summarize two typical types of transplanting operations in Fig.~\ref{fig:problem}(right). The core technique is to learn an adapter to connect a task module in the transplant net and a category module from another network.

\begin{figure}[t]
\centering
\includegraphics[width=0.86\linewidth]{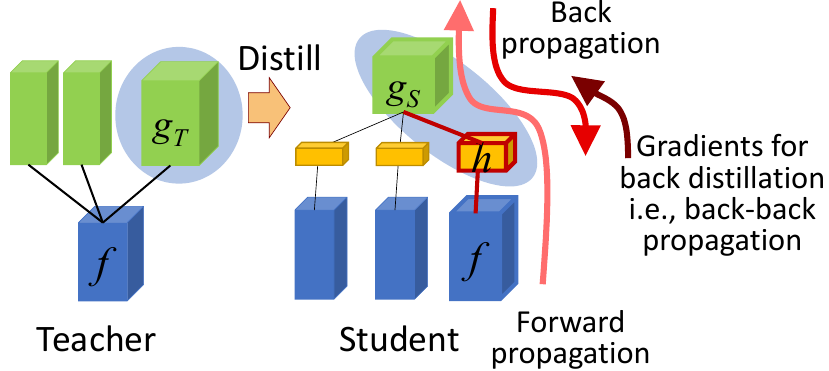}
\vspace{-10pt}
\caption{Overview. (left) Given a teacher net and a student net, we aim to learn an adapter $h$ via distillation. We transplant the category module $f$ to the student net by using $h$ to connect $f$ and {\small$g_{S}$}, in order to enable the task module {\small$g_{S}$} to deal with the new category $f$.\vspace{-15pt}}
\label{fig:task}
\end{figure}

The elementary transplanting operation is shown in Fig.~\ref{fig:task}. We are given a transplant net with a task module $g_{S}$ that is learned to accomplish a certain task for many categories, except for the category $c$. We hope the task module $g_{S}$ to deal with the new category $c$, so we need another network (namely, a \textit{teacher net}) with a category module $f$ and a task module $g_{T}$. The teacher net is pre-trained for the same task on the category $c$. We may (or may not) have a few training annotations of category $c$ for the task. Our goal is to transplant the category module $f$ in the teacher net to the transplant net.

Note that we just learn a small adapter module to connect $f$ to $g_{S}$. We do \textbf{not} fine-tune $f$ and $g_{S}$ during the transplanting process to avoid damaging their generality.

However, learning adapters but fixing parameters of category and task modules proposes specific challenges to deep-learning algorithms. Therefore, in this study, we proposed a new algorithm, namely \textit{back distillation}, to overcome these challenges. The back-distillation algorithm uses the cascaded modules of the adapter and $g_{S}$ to mimic upper layers of the pre-trained teacher net. This algorithm requires the transplant net to have similar gradients/Jacobian with the teacher net \emph{w.r.t.} $f$'s output features for distillation. In experiments, our back-distillation method without any training samples even outperformed the baseline with 100 training samples (see Table~\ref{tab:exp1}(left)).
\vspace{-5pt}

\subsubsection{Difference to previous knowledge transferring}

The proposed network transplanting is close to the spirit of continual learning (or lifelong learning). As an exploratory research, we summarize our essential differences from traditional studies in Table~\ref{tab:diff}.

\noindent
$\bullet\;$\textbf{Modular interpretability $\rightarrow$ more controllability:} Besides the discrimination power, the interpretability is another important property of a neural network. Our transplant net clarifies the functional meaning of each intermediate network module, which makes the knowledge-transferring process more controllable.

\noindent
$\bullet\;$\textbf{Bottleneck of transferring upper modules:} Most deep-learning strategies are not suitable to directly transfer pre-trained upper modules. Network transplanting just allows us to modify the lower adapter, when we transfer a pre-trained upper task module $g_{S}$ to a new category. It is not permitted to modify the upper module. However, it is difficult to optimize a lower adapter if the upper $g_{S}$ is fixed.

\noindent
$\bullet\;$\textbf{Catastrophic forgetting:} Continually learning new jobs without hurting the performance of old jobs is a key issue for continual learning~\cite{ProgressiveNN,continualLearning}. Our method exclusively learns the adapter to physically prevent the learning of new categories from changing existing modules.
\vspace{-5pt}

We can summarize contributions of this study as follows. (i) We propose a new deep-learning method, network transplanting with a few or even without additional training annotations, which can be considered as a theoretical solution to three issues in Section~\ref{sec:universal}. (ii) We develop an optimization algorithm, \emph{i.e.} back-distillation, to overcome specific challenges of network transplanting. (iii) Preliminary experiments proved the effectiveness of our method. Our method significantly outperformed baselines.

\section{Algorithm of network transplanting}

\textbf{Overview:} As shown in Fig.~\ref{fig:task}, we are given a teacher net for a single or multiple tasks \emph{w.r.t.} a certain category $c$. Let the category module $f$ in the bottom of the teacher net have $m$ layers, and it connects a specific task module {\small$g_{T}$} in upper layers. We are also given a transplant net with a generic task module {\small$g_{S}$}, which has been learned for multiple categories except for the category $c$.

The initial transplant net with a task module {\small$g_{S}$} (before transplanting) can be learned via traditional scenario of learning from samples of some categories. We can roughly regard {\small$g_{S}$} to encode generic representations for the task. Similarly, the category module $f$ extracts generic features for multiple tasks. Thus, we do not fine-tune {\small$g_{S}$} or $f$ to avoid decreasing their generality.

Our goal is to transplant $f$ to {\small$g_{S}$} by learning an adapter $h$ with parameters $\theta_{h}$, so that the task module {\small$g_{S}$} can deal with the new category module $f$.

The basic idea of network transplanting is that \textit{we use the cascaded modules of $h$ and {\small$g_{S}$} to mimic the specific task module {\small$g_{T}$} in the teacher net.} We call the transplant net a \textit{student net}. Let $x$ denote the output feature of the category module $f$ given an image $I$, \emph{i.e.} {\small$x=f(I)$}. {\small$y_{T}$} and {\small$y_{S}$} are given as outputs of the teacher net and the student net, respectively. Thus, network transplanting can be described as
\begin{small}
\vspace{-2pt}
\begin{equation}
\!y_{T}\!=\!g_{T}(x),\; y_{S}\!=\!g_{S}(h(x)), \; y_{T}\!\approx\!y_{S} \;\Rightarrow\; g_{S}(h(\cdot))\!\approx\!g_{T}(\cdot)\!\!\!
\vspace{-10pt}
\end{equation}
\end{small}

\subsection{Problem of space projection \& back-distillation}
\label{sec:challenge}

It is a challenge to let an adapter $h$ project the output feature space of $f$ properly to the input feature space of {\small$g_{S}$}. The information bottleneck theory shows that a network selectively \textit{forgets} certain space of middle-layer features and gradually focuses on discriminative features during the learning process. Thus, both the output of $f$ and the input of {\small$g_{S}$} have vast \textit{forgotten space}. Features in the forgotten input space of {\small$g_{S}$} cannot pass most feature information through ReLU layers in {\small$g_{S}$} and reach {\small$y_{S}$}. The forgotten output space of $f$ is referred to the space that does not contains $f$'s output features.

Vast forgotten feature spaces significantly boost difficulties of learning. Since valid input features of {\small$g_{S}$} usually lie in low-dimensional manifolds, most features of the adapter fall inside the forgotten space. \emph{I.e.} $g_{S}$ will not pass most information of input features to network outputs. Consequently, the adapter will not receive informative gradients of the loss for learning.

To learn good projections, we propose to force the gradient (also known as attention, Jacobian) of the student net to approximate that of the teacher, which is a necessary condition of {\small$g_{S}(h(\cdot))\!\approx\!g_{T}(\cdot)$}.
\begin{small}
\vspace{-2pt}
\begin{equation}
g_{S}(h(\cdot))\approx g_{T}(\cdot) \;\;\Longrightarrow\;\; \forall J(\cdot),\;\;\frac{\partial J(y_{S})}{\partial x}\propto \frac{\partial J(y_{T})}{\partial x}
\label{eqn:equal}
\vspace{-2pt}
\end{equation}
\end{small}
where $J(\cdot)$ is an arbitrary function of $y$ that outputs a scalar. $\theta_{h}$ denote parameters of the adapter $h$. Therefore, we use the following distillation loss for \textit{back-distillation}:
\begin{small}
\vspace{-2pt}
\begin{equation}
\!\underset{\theta_{h}}{\min}\,Loss,\;\; Loss=\mathcal{L}(y_{S},y^{*})+\lambda\cdot\Vert\alpha\frac{\partial J(y_{S})}{\partial x}\!-\!\frac{\partial J(y_{T})}{\partial x}\Vert^2\!\!
\vspace{-2pt}
\end{equation}
\end{small}
where {\small$\mathcal{L}(y_{S},y^{*})$} is the task loss of the student net; {\small$y^{*}$} denotes the ground-truth label; $\alpha$ is a scaling scalar. This formulation is similar to the Jacobian distillation in \cite{distillJacobian}. We omit {\small$\mathcal{L}(y_{S},y^{*})$}, if we learn the adapter without additional training labels.

\subsection{Learning via back distillation}
\label{sec:backDistill}

It is difficult for most recent techniques, including those for Jacobian distillation, to directly optimize the above back-distillation loss. To overcome the optimization problem, we need to make gradients of $J$ agnostic with regard to feature maps. Thus, we propose two pseudo-gradients {\small$D'_{S},D'_{T}$} to replace {\small$D_{S},D_{T}$} in the loss, respectively. The pseudo-gradients {\small$D'_{S},D'_{T}$} follow the paradigm in Eqn.~(\ref{eqn:DNew}).
\begin{small}
\vspace{-2pt}
\begin{subequations}
\begin{align}
D({\bf X},\theta_{h})&\xlongequal[]{\textrm{def}}\left.\frac{\partial J}{\partial x}\right|_{x=x^{(m)}}=G_{y}\frac{\partial y}{\partial x^{(n)}}\cdots\frac{\partial x^{(m+1)}}{\partial x^{(m)}}\label{eqn:D}\\
&=f'_{\textrm{conv}}\circ f'_{\textrm{relu}}\circ f_{\textrm{pool}}^{'\textrm{max}}\circ\cdots\circ f'_{\textrm{conv}}(G_{y})\nonumber\\
D'(\theta_{h})&\xlongequal[]{\textrm{def}}f'_{\textrm{conv}}\circ f'_{\textrm{dummy}}\circ f_{\textrm{pool}}^{'\textrm{avg}}\circ\cdots\circ f'_{\textrm{conv}}\left(G_{y'}\right)\label{eqn:DNew}
\vspace{-2pt}
\end{align}
\end{subequations}
\end{small}
where we define {\small$G_{y}=\frac{\partial J}{\partial y}$}. Just like in Eqn.~(\ref{eqn:equal}), we assume {\small$g_{S}(h(\cdot))\approx g_{T}(\cdot)\Rightarrow D'_{S}\propto D'_{T}$}. {\small$f'_1\circ f'_2(\cdot)\xlongequal[]{\textrm{def}}f'_1(f'_2(\cdot))$}, each $f'$ is the derivative of the layer function $f$ for back-propagation. {\small${\bf X}$} denotes a set of feature maps of all middle layers, and {\small$x^{(m)}\in{\bf X}$} is the feature map of the $m$-th layer. In Eqn.~(\ref{eqn:DNew}), we revise layer-wise operations to the computation of gradients, in order to make gradients {\small$D'$} agnostic with regard to {\small${\bf X}$}.

In this way, we conduct the back-distillation algorithm by {\small$\min_{\theta_{h}}Loss\!=\!\mathcal{L}(y_{S},y^{*})\!+\!\lambda\Vert\alpha D'_{S}\!-\!D'_{T}\Vert^2$}. The distillation loss can be optimized by propagating gradients of gradient maps to the upper layers, and we consider this as \textit{back-back-propagation}. Tables~\ref{tab:exp1} and \ref{tab:exp23} have exhibited the superior performance of the back-back-propagation.

\begin{table*}[th]
\begin{center}
\resizebox{0.95\linewidth}{!}{
\begin{tabular}{c|cl|rrrrr|r|c|cl|rrrrr|r|}
\cline{2-9}\cline{11-18}
\multirow{11}{*}{\begin{sideways}{Insert one conv-layer}\end{sideways}}&
\multicolumn{2}{l|}{\footnotesize{\# of samples}} & cat & cow & dog & horse & sheep & {\bf Avg.} &
\multirow{11}{*}{\begin{sideways}{Insert three conv-layers}\end{sideways}}&
\multicolumn{2}{l|}{\footnotesize{\# of samples}} & cat & cow & dog & horse & sheep & {\bf Avg.}\\
\cline{2-9}\cline{11-18}
&\multirow{2}{*}{100}&{direct-learn} & 12.89
& 3.09
& 12.89
& 10.82
& 9.28
& 9.79
&&
\multirow{2}{*}{100}&{direct-learn} & 9.28
& 6.70
& 12.37
& 11.34
& 3.61
& 8.66\\
&&{back-distill} & {\bf1.55}
& {\bf0.52}
& {\bf3.61}
& {\bf1.55}
& {\bf1.03}
& {\bf1.65}
&&
&{back-distill} & {\bf1.03}
& {\bf2.58}
& {\bf4.12}
& {\bf1.55}
& {\bf2.58}
& {\bf2.37}\\
\cline{2-9}\cline{11-18}
&\multirow{2}{*}{50}&{direct-learn} & 13.92
& 15.98
& 12.37
& 16.49
& 15.46
& 14.84
&&
\multirow{2}{*}{50}&{direct-learn} & 14.43
& 13.92
& 15.46
& 8.76
& 7.22
& 11.96\\
&&{back-distill} & {\bf1.55}
& {\bf0.52}
& {\bf3.61}
& {\bf1.55}
& {\bf1.03}
& {\bf1.65}
&&
&{back-distill} & {\bf3.09}
& {\bf3.09}
& {\bf4.12}
& {\bf2.06}
& {\bf4.64}
& {\bf3.40}\\
\cline{2-9}\cline{11-18}
&\multirow{2}{*}{20}&{direct-learn} & 16.49
& 26.80
& 28.35
& 32.47
& 25.77
& 25.98
&&
\multirow{2}{*}{20}&{direct-learn} & 22.16
& 25.77
& 32.99
& 22.68
& 22.16
& 25.15\\
&&{back-distill} & {\bf1.55}
& {\bf0.52}
& {\bf3.09}
& {\bf1.55}
& {\bf1.03}
& {\bf1.55}
&&
&{back-distill} & {\bf7.22}
& {\bf6.70}
& {\bf7.22}
& {\bf2.58}
& {\bf5.15}
& {\bf5.77}\\
\cline{2-9}\cline{11-18}
&\multirow{2}{*}{10}&{direct-learn} & 39.18
& 39.18
& 35.05
& 41.75
& 38.66
& 38.76
&&
\multirow{2}{*}{10}&{direct-learn} & 36.08
& 32.99
& 31.96
& 34.54
& 34.02
& 33.92\\
&&{back-distill} & {\bf1.55}
& {\bf0.52}
& {\bf3.61}
& {\bf1.55}
& {\bf1.03}
& {\bf1.65}
&&
&{back-distill} & {\bf8.25}
& {\bf15.46}
& {\bf10.31}
& {\bf13.92}
& {\bf10.31}
& {\bf11.65}\\
\cline{2-9}\cline{11-18}
&\multirow{2}{*}{\textcolor{red}{\bf0}}&{direct-learn} & -- & -- & -- & -- & -- & --
&&
\multirow{2}{*}{\textcolor{red}{\bf0}}&{direct-learn} & -- & -- & -- & -- & -- & --\\
&&{back-distill} & {\bf1.55}
& {\bf0.52}
& {\bf4.12}
& {\bf1.55}
& {\bf1.03}
& {\bf1.75}
&&
&{back-distill} & {\bf50.00}
& {\bf50.00}
& {\bf50.00}
& {\bf49.48}
& {\bf50.00}
& {\bf49.90}\\
\cline{2-9}\cline{11-18}
\end{tabular}}
\vspace{-10pt}
\caption{Error rates of classification when we insert $n$ conv-layers with ReLU layers to a CNN as the adapter. {\small$n\in\{1,3\}$}. The last row shows the performance of network transplanting without training samples.}
\label{tab:exp1}
\resizebox{0.9\linewidth}{!}{\begin{tabular}{cl|rrrrr||cl|rrrrr}
\hline
\multicolumn{2}{l|}{\footnotesize{$\!\!\!\!\!$\# of samples}} & cat & cow & horse & sheep & {\bf Avg.} & \multicolumn{2}{l|}{\footnotesize{\# of samples}} & cat & cow & horse & sheep & {\bf Avg.}\\
\hline
\multirow{3}{*}{100}&{direct-learn}& 76.54
& 74.60
& 81.00
& 78.37
& 77.63
&\multirow{3}{*}{20}&{direct-learn}& 71.13
& 74.82
& 76.83
& 77.81
& 75.15\\
&{distill} &74.65
& 80.18
& 78.05
& 80.50
& 78.35
&&{distill} &71.17
& 74.82
& 76.05
& 78.10
& 75.04\\
&{back-distill} & {\bf85.17}
& {\bf90.04}
& {\bf90.13}
& {\bf86.53}
& {\bf87.97}
&&{back-distill} & {\bf84.03}
& {\bf88.37}
& {\bf89.22}
& {\bf85.01}
& {\bf86.66}\\
\hline
\multirow{3}{*}{50}&{direct-learn}& 71.30
& 74.76
& 76.83
& 78.47
& 75.34
&\multirow{3}{*}{10}&{direct-learn}& 70.46
& 74.74
& 76.49
& 78.25
& 74.99\\
&{distill} &68.32
& 76.50
& 78.58
& 80.62
& 76.01
&&{distill} &70.47
& 74.74
& 76.83
& 78.32
& 75.09\\
&{back-distill} & {\bf83.14}
& {\bf90.02}
& {\bf90.46}
& {\bf85.58}
& {\bf87.30}
&&{back-distill} & {\bf82.32}
& {\bf89.49}
& {\bf85.97}
& {\bf83.50}
& {\bf85.32}\\
\hline
\end{tabular}}
\vspace{-10pt}
\caption{Pixel accuracy of object segmentation when we transplanted the segmentation module from a \textit{dog} network to the network of the target category in Experiment 3. The adapter contained a conv-layer and a ReLU layer.\vspace{-15pt}}
\label{tab:exp23}
\end{center}
\end{table*}

\begin{figure}
\centering
\includegraphics[width=0.99\linewidth]{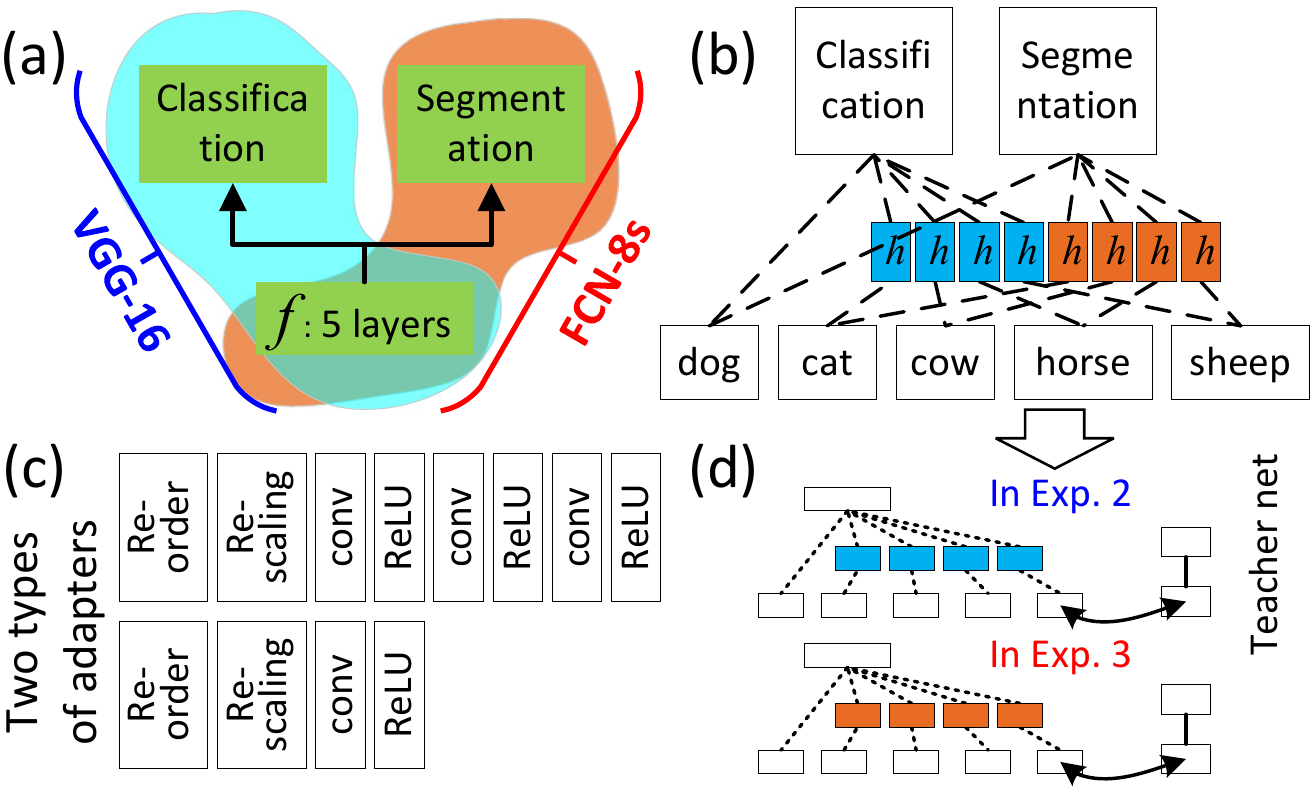}
\vspace{-10pt}
\caption{(a) Teacher net; (b) transplant net; (c) two types of adapters; (d) two sequences of transplanting operations.}
\label{fig:exp23}
\vspace{-12pt}
\end{figure}

\section{Experiments}

To simplify the story, we limit our attention to testing network transplanting operations. We do not discuss other related operations, \emph{e.g.} the fine-tuning of category and task modules and the case in Fig.~\ref{fig:problem}(a), which can be solved via traditional learning strategies.

We designed three experiments to evaluate the proposed method. In Experiment~1, we learned toy transplant nets by inserting adapters between middle layers of pre-trained CNNs. Then, Experiments~2 and 3 were designed considering the real application of learning a transplant net with two task modules (\emph{i.e.} modules for object classification and segmentation) and multiple category modules. As shown in Fig.~\ref{fig:exp23}(b,d), we can divide the entire network-transplanting procedure into an operation sequence of transplanting category modules to the classification module and another operation sequence of transplanting category modules to the segmentation module. Therefore, we separately conducted the two sequences of transplanting operations in Experiments~2 and 3 for more convincing results.

We compared our back-distillation method (namely \textit{back-distill}) with two baselines. All baselines exclusively learned the adapter without fine-tuning the task module for fair comparisons. The first baseline only optimized the task loss {\small$\mathcal{L}(y_{S},y^{*})$} without distillation, namely \textit{direct-learn}. The second baseline is the traditional distillation, namely \textit{distill}, where the distillation loss is {\small$CrossEntropy(y_{S},y_{T})$}. The distillation was applied to outputs of task modules {\small$g_{S}$} and {\small$g_{T}$}, because except for outputs, other layers in {\small$g_{S}$} and {\small$g_{T}$} did not produce features on similar manifolds. We tested the \textit{distill} method in object segmentation, because unlike single-category classification, segmentation outputs had correlations between soft output labels.

\subsection{Exp. 1: Adding adapters to pre-trained CNNs}

In this experiment, we conducted a toy test, \emph{i.e.} inserting and learning an adapter between a category module and a task module to test network transplanting. Here, we only considered networks with VGG-16 structures for single-category classification.

In Table \ref{tab:exp1}, we compared our \textit{back-distill} method with the \textit{direct-learn} baseline when we inserted an adapter with a single conv-layer and when we inserted an adapter with three conv-layers. Table~\ref{tab:exp1}(left) shows that compared to the $9.79\%$--$38.76\%$ error rates of the \textit{direct-learn} baseline, our \textit{back-distill} method yielded a significant lower classification error ($1.55\%$--$1.75\%$). \textit{Even without any training samples, our method still outperformed the \textit{direct-learn} method with 100 training samples.}

\subsection{Exp. 3: Operation sequences of transplanting category modules to the segmentation module}

In this experiment, we evaluated the performance of transplanting category modules to the segmentation module. Five FCNs were strongly supervised using all training samples for single-category segmentation. We considered the segmentation module of the dog as a generic one. We transplanted category modules of other four mammal categories to this task module.

Table~\ref{tab:exp23} compares pixel-level segmentation accuracy between our method and the \textit{direct-learn} baseline. We tested our method with a few (10--100) training samples. Our method exhibited 10\%--12\% higher accuracy than the \textit{direct-learn} baseline.

\section{Conclusions and discussion}

In this paper, we focused on a new task, \emph{i.e.} merging pre-trained teacher nets into a generic, modular transplant net with a few or even without training annotations. We discussed the importance and core challenges of this task. We developed the back-distillation algorithm as a theoretical solution to the challenging space-projection problem. The back-distillation strategy significantly decreases the demand for training samples. Experimental results demonstrated the superior efficiency of our method. Our method without any training samples even outperformed the baseline with 100 training samples, as shown in Table~\ref{tab:exp1}(left).

{\small
\bibliographystyle{aaai}
\bibliography{TheBib}
}

\end{document}